\documentclass[letterpaper, 10pt, conference]{ieeeconf}   

\IEEEoverridecommandlockouts               
\usepackage{cite}
\usepackage{color,xcolor}
\usepackage{graphicx}
\usepackage{pdfpages}
\usepackage{subfigure}
\usepackage{tabulary}
\usepackage{balance}

\usepackage{amsmath}
\usepackage{amssymb}
\usepackage{float}
\usepackage{soul}
\usepackage{booktabs} 
\usepackage{multirow}  
\usepackage{hyperref} 
\usepackage{array}  
\usepackage{diagbox}

\usepackage{mathtools}
\graphicspath{{pics/}}
\usepackage{bm}
\usepackage{verbatim}

\renewcommand{\baselinestretch}{0.96} 

\begin{document}
	
	\title{Two-Stage Learning of Highly Dynamic Motions \\with Rigid and Articulated Soft Quadrupeds}
	
	\author{Francecso Vezzi\textsuperscript{1,$\star$}, Jiatao Ding\textsuperscript{1,$\star$,*}, Antonin Raffin\textsuperscript{2}, 
		Jens Kober\textsuperscript{1}, Cosimo Della Santina\textsuperscript{1,2}
		\thanks{This work is supported by the EU project 101016970 NI. 
			$^1$Authors are with the Department of Cognitive Robotics, Delft University of Technology, Building 34, Mekelweg 2, 2628 CD Delft, Netherlands (f.vezzi.96@gmail.com, J.Ding-2@tudelft.nl, J.Kober@tudelft.nl, C.DellaSantina@tudelft.nl). 
			%
			$^2$Authors are with the Institute of Robotics and Mechatronics, German Aerospace Center (DLR), 82234 Wessling, Germany (Antonin.Raffin@dlr.de).
			$^\star$ These authors contributed equally to this work. 
			$^*$ Jiatao Ding is the corresponding author. }
	}
	
	\maketitle
	
	\begin{abstract}
		
		Controlled execution of dynamic motions in quadrupedal robots, especially those with articulated soft bodies, presents a unique set of challenges that traditional methods struggle to address efficiently. In this study, we tackle these issues by relying on a simple yet effective two-stage learning framework to generate dynamic motions for quadrupedal robots. First, a gradient-free evolution strategy is employed to discover simply represented control policies, eliminating the need for a predefined reference motion. Then, we refine these policies using deep reinforcement learning. Our approach enables the acquisition of complex motions like pronking and back-flipping, effectively from scratch. Additionally, our method simplifies the traditionally labour-intensive task of reward shaping, boosting the efficiency of the learning process. Importantly, our framework proves particularly effective for articulated soft quadrupeds, whose inherent compliance and adaptability make them ideal for dynamic tasks but also introduce unique control challenges. 

	\end{abstract}

	\IEEEpeerreviewmaketitle
	
	\section{Introduction} \label{introduction}
	
	Quadrupedal robots \cite{hutter2016anymal,katz2019mini,unitreego1} 
	promise to address real-world challenges due to their potential advantages in terms of mobility and versatility. To improve the performance and extend mission time \cite{angelini2023robotic}, elastic and soft elements \cite{della2021soft} have been applied to quadrupedal robots in recent works, such as \cite{badri2022birdbot,bjelonic2023learning,arm2019spacebok,raffin2022learning}. Although stable locomotion can be achieved by off-the-shelf solvers, 
	accomplishing highly dynamic motion 
	is still an open challenge due to the under-actuated, hybrid, nonlinear dynamics. 

	Acrobatic motions have been achieved with quadrupedal robots by virtue of trajectory optimization (TO) and model predictive control (MPC), including jumping \cite{nguyen2019optimized,nguyen2022continuous,song2022optimal} and back-flip \cite{ding2021representation,garcia2021time}. However, these approaches are computationally intensive and require accurate models. To guarantee the solution when solving the TO problems, simplifications in problem formulations are often needed, resulting in a conservative solution. 

	Reinforcement learning (RL) has emerged as a promising alternative to model-based controllers for achieving dynamic motions. Although normal gaits like trotting can be easily learned from scratch, learning acrobatic motions needs extra setup, such as tedious reward shaping \cite{hwangbo2019learning,fu2021minimizing,rudin2022learning} and careful curriculum design \cite{tang2021learning,margolis2022rapid,iscen2021learning,chen2022learning}, due to the highly nonlinear dynamics and reward sparsity. To address this problem, the current RL frameworks assume the availability of prior knowledge on how to solve the task in the form of a reference motion/trajectory \cite{margolis2021learning,fuchioka2023opt,bogdanovic2021model,li2022versatile,li2023learning}, a pre-existing controller \cite{smith2023learning,yang2023continuous} or both \cite{yang2023cajun,li2023robust}.  
	%
	This reliance on expert knowledge limits the applicability of these methods as it may not be readily available and possibly biases the learning towards sub-optimal policies.
	
	Apart from motion generation for rigid robots, many works have argued that introducing elasticity can yield better performance \cite{kashiri2018overview}, which, however, brings extra control challenges. Up to now, very limited work has been done on generating highly dynamic quadrupedal motions exploiting parallel elasticity. RL-based works such as \cite{bjelonic2023learning} and \cite{raffin2022learning} employ passive elasticity for quadrupedal locomotion but do not report such acrobatic motions as done in this work. The only exception is SpaceBok \cite{arm2019spacebok}, which is, however, specifically designed to work in a low-gravity environment.
	
\begin{figure*}
		\centering
		\setlength{\belowcaptionskip}{-0.cm}
		\includegraphics[width=2\columnwidth]{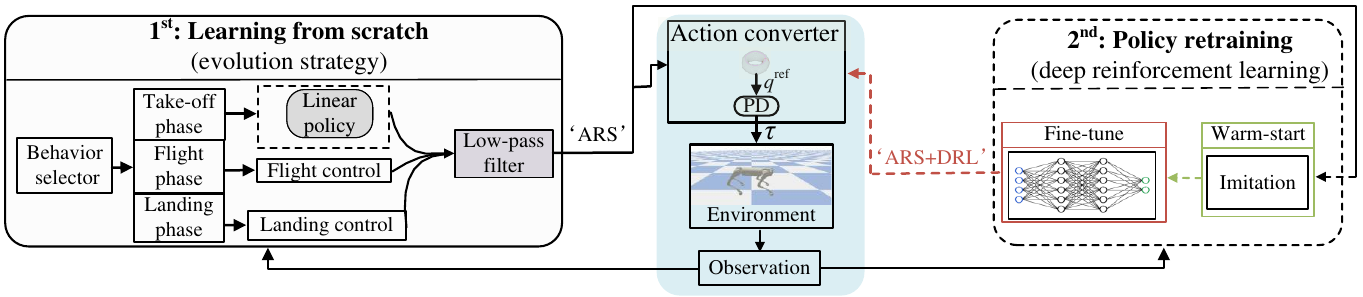}
		\vspace{-0.3cm}
		\caption{{Two-stage learning procedure for the acrobatic motion with a flight phase. Without defining reference motions, the (rigid and soft) quadrupedal robot realizes (a) jumping in place and jumping forward, (b) pronking and (c) back-flip}. `\texttt{ARS}' and `\texttt{ARS+DRL}' separately represent the linear policies generated by first-stage ES and the refined policy after the second-stage retraining.}
		\label{fig:demonstration_snapshots}
		\vspace{-0.3cm}
	\end{figure*}

	To overcome these limitations, we propose here a two-stage strategy for learning highly dynamic motions from scratch by combining RL with an evolution strategy (ES), as illustrated in Fig.~\ref{fig:demonstration_snapshots}. To start with, an evolution stage is employed to directly learn joint commands meeting the task demands, without requiring an expert demonstration defined in advance. In this stage, we use the augmented random search (\texttt{ARS}) \cite{mania2018simple}  algorithm to learn a linear policy. Then, we transfer the learned policy to a more expressive neural network via deep reinforcement learning (DRL) \cite{franccois2018introduction}. We use proximal policy optimization (\texttt{PPO}) \cite{schulman2017proximal} to refine the policy that is warm-started by mimicking the learned motions in the first stage. Note that a similar two-stage method containing ES and DRL has been proposed in \cite{shi2022reinforcement} for quadrupedal locomotion. However, the ES, specifically the natural evolution strategies algorithm \cite{salimans2017evolution},  was used there to search the parameters for a center pattern generator that generates reference trajectories for the second-stage RL, and no dynamic motions like ours were reported.
	
\begin{figure*}
		\centering
		\setlength{\belowcaptionskip}{-0.cm}
		\includegraphics[width=2\columnwidth]{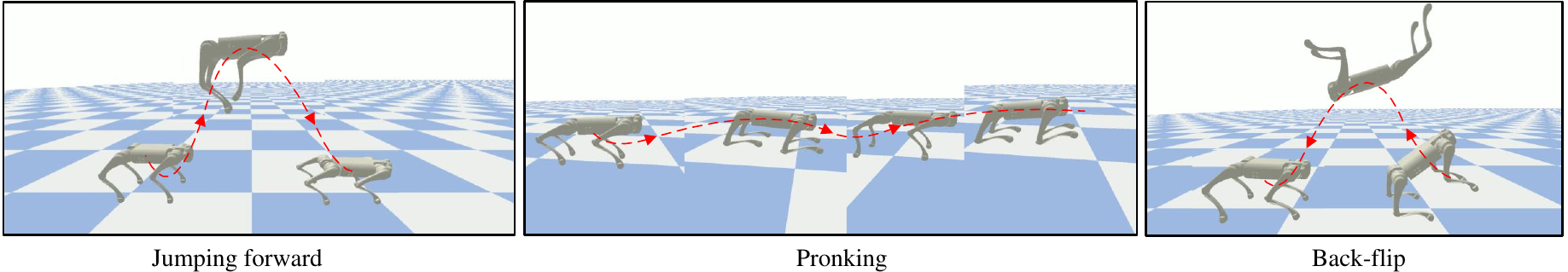}
		\vspace{-0.3cm}
		\caption{Examples of highly dynamic motions that the robot could learn using the proposed strategy. Red curves plot the CoM movements where the arrows point to the movement directions.}
		\label{fig:results_snapshots}
		\vspace{-0.6cm}
	\end{figure*}

	We validate\footnote{All the videos can be found at: \url{https://www.youtube.com/watch?v=liIeYU71a5w}} the approach by rich simulations when learning versatile acrobatic motions, including jumping, pronking and back-flip, see Fig.~\ref{fig:results_snapshots}. Furthermore, we apply the approach to both a rigid and an articulated soft quadruped. We demonstrate that the proposed architecture enables exploiting parallel elasticity in achieving highly dynamic motions. Taking the jumping task as an example, the articulated soft quadrupedal robot could jump 15.4\% higher in the in-placed jumping task, and jump 23.1\% further in the forward jumping.
	
	\section{Methodology} \label{prelinimaries}
	We introduce basic notation in Section~\ref{sec:notaion} and the two-stage procedure in a general fashion in Section~\ref{two-stage-learning}. Acrobatic-specific details follow in Section~\ref{parallel_springs} - yielding the general architecture shown in Fig.~\ref{fig:demonstration_snapshots}. Task-specific details will be provided later in Section~\ref{DRP-from-scratch}.
	%
	%
	\subsection{RL notation}\label{sec:notaion}
	The RL problem is formulated as a {Markov decision process (MDP)}, described by the tuple  \resizebox{0.3\hsize}{!}{$\mathcal{M} = \langle \mathcal{S}, \mathcal{A}, \mathcal{P}, \mathcal{R}, \gamma \rangle $ }where \(\mathcal{S}\) is the state space, \(\mathcal{A}\) is the action space, \(\mathcal{P}\) is the transition probability, \(\mathcal{R}\) is the reward distribution, and \(\gamma\) is the discount factor. 
	The states \((\bm{s}_k \in \mathcal{S})\), actions \((\bm{a}_k \in \mathcal{A})\), and rewards \((r(\bm{s}_k, \bm{a}_k) \in \mathcal{R})\) in an episode of length \(T\) constitute a rollout of the policy. In each rollout, the cumulative return is  \resizebox{0.4\hsize}{!}{$G_T = \sum_{k = 1}^{T} \gamma^{k - 1} r(\bm{s}_k, \bm{a}_k)$}. In the following, we will assume that the whole state is observable, and thus we will refer to elements in \(\mathcal{S}\) as observations.
	The goal of RL is then to find an approximation of the optimal policy ($\pi^{\star}$) such that the agent achieves the maximal expected return, i.e., $\pi^{\star} = \arg\max\limits_{ {\pi}} \mathbb{E}\left[G_T \mid \pi\right]$. 
	
	\subsection{Two-stage learning}\label{two-stage-learning}
	\subsubsection{First step-learning a simple policy from scratch via evolution strategy}\label{first_stagge_metho}
	Despite DRL's success in sequential decision-making, it struggles with sparse-reward tasks
	\cite{majid2023deep}. To address this, we use a gradient-free ES scheme, specifically \texttt{ARS}, in the first stage to treat the RL tasks as black-box optimization \cite{hansen2015evolution},
	%
	where the linear policy is adopted\footnote{The work in \cite{mania2018simple} argued that the linear policy is simple but enough to obtain competitive results for RL tasks. A comparison with other representations on the jumping task is found in Section~\ref{comparison_representation}. }
	\begin{equation}\small
		\pi_{\texttt{ars}}(\bm{s}_k) =\mathbf{W}\,\bm{s}_k +\bm{b},
		\label{ars_line_policy}
	\end{equation}
	where $\mathbf{W}: \mathcal{S} \rightarrow \mathcal{A}$ is a linear application, and $\bm{b}\in\mathcal{A}$. 
	\subsubsection{Second step-refining and retraining via DRL}\label{methodology_second_stage}
	We then resort to the more expressive deep neural network (DNN) to refine the policy. Using the DRL method, specifically \texttt{PPO}, we achieve the nonlinear expression
	\begin{equation}\small\label{eq:PPO}
		\pi_{\texttt{PPO}}(\bm{a}_k|\bm{s}_k; {\bm{\theta}}) = \mathrm{NN}(\bm{s}_k; \bm{\theta}), 
	\end{equation}   
	where $\mathrm{NN}:\mathcal{S}\times\mathbb{R}^n \rightarrow \Delta(\mathcal{A})$ is a policy function that maps states to a probability distribution on the action space, and $\bm{\theta} \in \mathbb{R}^n$ is the vector of the network parameters.

	At this stage, we start training a DNN by replicating the observation-action pair provided by the first-stage policy, using the imitative reward function 
	$r^{\text{ini}} = w_\mathrm{a} \text{exp}^{(-w_\mathrm{b} ||\bm{a} - \bm{a}^{\texttt{ars}} ||^2)}$, with $\bm{a}, \bm{a}^{\texttt{ars}} \in \mathcal{A}$ respectively being the output of the DNN and the first-stage \texttt{ARS} given an observation, and $w_\mathrm{a}, w_\mathrm{b} \in \mathbb{R}$ being the hyperparameters.
	Afterwards, we fine-tune the DNN using the task-specific reward ${}^2r$, 
	which we detail in section \ref{DRP-from-scratch}.

	\subsection{Learning acrobatic motions for quadrupeds} \label{parallel_springs}
	
\begin{figure}
		\centering
		\setlength{\belowcaptionskip}{-0.cm}
		\includegraphics[width=\columnwidth]{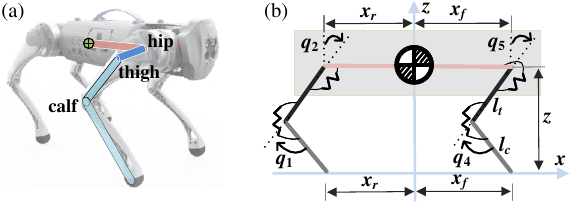}
		\vspace{-0.4cm}
		\caption{Quadrupedal robot Go1 (a) and its PEA arrangement (b). In the homing pose, we have initial height $z_0=0.32$m. We call $q_1,q_4$ the calf angles, $q_2,q_5$ the thigh angles, and $q_3,q_6$ the hip angles.}
		\label{fig:quadrupedal_slip}
		\vspace{-0.4cm}
	\end{figure}

	%
	We define an acrobatic motion as a dynamic movement involving three (possibly repeated) phases: (i) take-off, (ii) flight, (iii) landing.
	%
	On top of what we discussed in the previous section, we introduce a \textit{behavior selector} mechanism. 
	This is a state machine that transitions through the three phases discussed above, characterizing every acrobatic motion.  The first phase is the take-off, which uses \eqref{ars_line_policy}. The second is the flight phase, which generates a constant action equal to the homing pose in Fig.~\ref{fig:quadrupedal_slip}(b). The third phase passes the same pose through a proportional-derivative (PD) controller thus reducing the effective joint stiffness. This is sufficient to ensure a compliant landing.
	%
	%
	The guarding conditions of the flight phase and landing phase separately are \( F_{\text{contact}} = 0 \), and $\dot{h} < 0$ \texttt{and} $  F_{\text{contact}} > 0$, where \( F_{\text{contact}} \) is the total contact force between the robot feet and the ground, and \( \dot{h} \) is the vertical velocity.
	We also add a low-pass filter to the action $a$ before feeding it into the environment as \texttt{ARS} can result in non-smooth control signals. 
	%
	%
	We remove these additions in the second stage of learning, relying on the capability of the $\textit{NN}$ in \eqref{eq:PPO} to learn the smooth actions. 
	
	

	\section{Learning task formulation 
	} \label{DRP-from-scratch}
	
	This section defines the key elements of the RL tuple for learning acrobatic motions. We specifically focus on jumping, pronking, and back-flip (see Fig. \ref{fig:results_snapshots}). Note that if not differently specified, the observation and action spaces are the same during the two learning stages and the three tasks.
	
	We conduct simulation experiments using the Go1 robot from UniTree and its soft variant featuring parallel elastic actuation, as detailed in Appendix~\ref{parallel_elsticity}. Both models are depicted in Fig. \ref{fig:quadrupedal_slip}. Hence, when discussing specifics, we'll assume these are the robots in question, although our architecture is broadly applicable to any quadrupedal system. Moreover, it is worth pointing out here that we will focus on Sagittal motions - although the environment is fully 3D.
	
	\subsection{Observation/state} 
	These include the base (i.e., trunk) height ($h\in \mathbb{R}$), vertical velocity ($\dot{h}\in \mathbb{R}$), and the pitch angle ($\theta\in \mathbb{R}$). Furthermore, the average\footnote{More precisely, $q_i = (q_i^{\mathrm{left}} + q_i^{\mathrm{right}})/2$, where $q_i^{\mathrm{left}},q_i^{\mathrm{right}} \in \mathbb{R}$ are the $i$-th joint according to the numbering in Fig. \ref{fig:quadrupedal_slip} on the left and right sides of the robot when cut by the Sagittal plane.} joint angles ($\bm{q}\in \mathbb{R}^6$) and velocities ($\dot{\bm{q}}\in \mathbb{R}^6$) are also added to the observations. Finally, we include a Boolean flag $b\in\{0,1\}$ that indicates whether the robot is in a flight phase or not. This observation is, however, only available to $\pi_{\mathrm{PPO}}$. The resulting observations vector is $\bm{s}_k = \{h, \dot{h}, \theta, \bm{q}, \dot{\bm{q}}, b \}$.  In simulations, the white noise is added to the observations to emulate the sensory noise.
	
	

	\subsection{Action space} Assuming symmetry of the robot and of the task between the right and left sides, we only learn the 3 joints in the front leg and 3 joints in the rear leg. 
	
	We convert the policy outputs (i.e., $\bm{a}_k \in [-1, 1]^6$) to joint angle ($\bm{q}^\text{ref}$) commands by a linear scaling defined such that the maximum actions corresponds to the joint limits. which we feed into a PD controller that generates motor torques as follows
	$\tau_i = k_\text{p} (q^\text{ref}_i - q_i) - k_\text{d}\dot{q}_i,$
	where $k_\text{p} \in \mathbb{R} $ and $k_\text{d} \in \mathbb{R}$ are the proportional and derivative gains, respectively.	
	
	The pipeline of converting action to motor torque command is summarized by the `Action converter' in Fig.~\ref{fig:demonstration_snapshots}.

	\subsection{Episode design} 
	In our settings, one episode is {terminated when the time runs out or early termination is triggered}. In particular, the early termination is triggered if one of the following is true
	\begin{itemize}
		\item the robot falls, which we detect by checking the height, i.e., $h<0.1$ m,
		\item the knees touch the ground, or links collide with each other,
		\item the trunk is parallel enough to the ground, 
		that we quantify by evaluating the scalar product between the local $z$-axis and the global one and testing that it is less than $0.85$.
	\end{itemize}
	The third condition is not present in the back-flip task.

	\subsection{Rewards for the two stages 
	}\label{reward_fine_tuning}
	
	\subsubsection{Notation}
	We concisely refer to the exponential kernel as follows
	\begin{equation}
		\resizebox{0.5\hsize}{!}{$
			g_l^t : \mathbb{R} \to \mathbb{R}, \quad g_l^t(x) \coloneqq  a_l^t e^{- b_l^t |x|}
			\label{eq:exper_kernel}
			$}
	\end{equation}
	with $a_l^t$ and $b_l^t$ being hyperparameters. 
	%
	%
	With this notation, the kernel function for different tasks is encapsulated inside the subscript and superscript, where the superscript ($t$) specifies the task, while the subscript ($l$) refers to the input variable. 
	Hereafter, we use $r$ to denote $r(\bm{s}_k,\bm{a}_k)$ for brevity

	\subsubsection{Reward shaping} 
	To define the reward function for each task, we first normalize and clip the maximal height ($\overline{h} \in \mathbb{R})$. For the jumping task, we also normalize and clip the maximal jumping distance ($\overline{d} \in \mathbb{R}$). That is
	\begin{equation}
		\resizebox{0.8\hsize}{!}{$
			\begin{aligned}
				h_n = \text{clip} \left\{0,  {\overline{h}}/{h_{\max}}, 1 \right\}, \quad \!\!
				d_n = \text{clip} \left\{0,  {\overline{d}}/{d_{\max}}, 1 \right\},
			\end{aligned}
			\label{normalization_xx}
			$}
	\end{equation}
	where $h_{\max}$ and $d_{\max}$ denote the maximal height and maximal jumping distance that can be achieved by the robot, which we tuned heuristically. 
	
	We go now more in detail, taking the jumping task as an example. Pronking and back-flip are similarly defined by making minor changes to the reward functions. More details can be found in Appendix~\ref{reward_all}. 
	
	\paragraph{Reward function for the first stage}\label{jumping_first_stage}	
	Since ES search can deal with the learning tasks with sparse rewards, alleviating the efforts for rewarding shaping, we assume that the robot does not get any reward during one episode until it ends. The other reason is that dense reward might lead to unwanted behavior, whereas sparse reward makes it simple to obtain the correct behavior. Thus, the reward for the first-stage \texttt{ARS} training (${}^1r^{\text{jump}}$) is straightforwardly defined as
	\begin{equation}
		\resizebox{0.8\hsize}{!}{$
			{}^1r^{\text{jump}} = \begin{cases}
				0 & \text{when episode is running}, \\
				r_{\text{bonus}}+ r_{\text{end}} & \text{when episode terminates}, 
			\end{cases}
			\label{fig:reward_function}
			$}
	\end{equation}	
	where the ending reward $r_{\text{end}}$ is computed by evaluating the state when the episode terminates, and the bonus reward $r_{\text{bonus}}$ is by evaluating the jumping performance.
	
	To achieve a jumping motion, we expect the robot to increase its normalized height ($h_n$) as much as possible so as to take off from the ground. Besides, to reduce the control efforts for posture adjustment in the air while minimizing the landing recovery efforts after touchdown, we penalize the maximal pitch angle ($\overline{\theta} \in \mathbb{R}$) the robot experiences during one episode. Finally, one additional term is added to regulate the normalized jumping distance ($d_n$). For jumping in place, this term minimizes the forward distance while, for jumping forward, this term encourages a large forward distance.

	\textbf{Reward for jumping in place} are defined as
	\begin{equation}\small
		\resizebox{0.8\hsize}{!}{$
			\begin{aligned}\small
				r_{\text{end}}^\text{jip}  &= 
				\left[g_{{\theta}}^\text{jip}(\overline{\theta}) + g_d^\text{jip}(d_n) + c_h^\text{jip} \right] h_n ,\\
				r_{\text{bonus}}^\text{jip} &= \begin{cases}
					+\, b^\text{jip} h_n & \text{not early termination}, \\
					-\left[ q^\text{jip} + m^\text{jip} h_n \right] & \text{otherwise}, \label{eq:r_prize-jip}
				\end{cases}
			\end{aligned}
			$}
	\end{equation}
	where $g_{\theta}^\text{jip}$ and $g_{d}^\text{jip}$ are computed using \eqref{eq:exper_kernel}. $c_h^\text{jip}$, $b^\text{jip}$, $q^\text{jip}$ and $m^\text{jip}$ are the positive hyperparameters.

	
	\textbf{Reward for jumping forward} are then defined as
	\begin{equation}\small
		\resizebox{0.9\hsize}{!}{$
			\begin{aligned}
				r_{\text{end}}^\text{jf}  &= 
				\left[g_{{\theta}}^\text{jf}(\overline{\theta}) + c_d^\text{jf} d_n  + c_h^\text{jf} \right] h_n , \label{eq:r_end-jf}\\
				r_{\text{bonus}}^\text{jf} &= \begin{cases}
					+\,  b^\text{jf} (h_n + d_n) & \text{no early termination}, \\
					-\left[ q^\text{jf} +  m^\text{jf} (h_n + d_n)   \right] & \text{otherwise},
				\end{cases}
			\end{aligned}
			$}
	\end{equation}
	where $g_{\theta}^\text{jf}$ is computed using \eqref{eq:exper_kernel}. $c_d^\text{jf}$, $c_h^\text{jf}$, $b^\text{jf}$, $q^\text{jf}$ and $m^\text{jf}$ are the positive hyperparameters.
	
	Note that the reward functions for jumping forward are quite similar to those for jumping in place. The only difference is that the jumping distance appears proportionally when computing $r_{\text{end}}^\text{jf}$ rather than inside the kernel function such that a longer jumping distance is encouraged. Besides, when computing the bonus for jumping forward ($r_{\text{bonus}}^\text{jf}$), we consider both the normalized maximal jumping distance $d_n$ and the normalized maximal height $h_n$.

	\paragraph{Reward function for the second stage} \label{learing_imitation}
	We discuss here the reward function used for refinement. The reader can refer to Section~\ref{methodology_second_stage} for the warm-starting reward $r^{\text{ini}}$.
	%
	Then, the dense refinement reward (${}^2r$) for retraining is 
	\begin{equation}
		{}^2r^{\text{jump}} = r_{h} + r_{c} + r_{d} + r_{s} + r_{\theta} + {}^2r_{\text{bonus}}.
		\label{eq:reward_second_stage}
	\end{equation}
	
	In \eqref{eq:reward_second_stage}, the term $r_h$ is defined as
	\begin{equation}
		\resizebox{0.5\hsize}{!}{$
			r_{h} = \begin{cases}
				a_h h & \text{if } h_{\min} \leq h \leq h_{\max}, \\
				0 & \text{otherwise},
			\end{cases}
			$}
	\end{equation} 
	with $a_h$ being a positive hyperparameter, which encourages the agent to not exceed $h_{\max}$ when jumping and avoids rewarding it when not jumping at all ($h < h_{\min}$). Note that here we use time-varying height $h$ instead of the normalized maximal height $h_n$.
	
	The term $ r_{c}$ is used to penalize the average contact force ($f  \in \mathbb{R}$) so as to encourage soft landing.
	\begin{equation}
		\resizebox{0.4\hsize}{!}{$
			r_{c} = \begin{cases}
				-a_c f & \text{if } f \geq f_{\min}, \\
				0 & \text{otherwise},
			\end{cases}
			$}
	\end{equation}
	where $f$ is computed by dividing total force ($F_\text{contact}$) by the number of contact legs, $f_{\min}$ is the lower threshold, and $a_c$ is a hyperparameter. 
	
	The term $r_{d}$ regulates jumping distance ($d \in \mathbb{R} $). Particularly, for jumping in place, we minimize the jumping distance whereas, for jumping forward, we maximize the jumping distance. To this end, we have
	\begin{equation}
		\begin{aligned}
			r_{d}^{\text{jip}} = g_{d}^{\text{jip}}(d), \quad
			r_{d}^{\text{jf}} = k_d^{\text{jf}} d. \\
		\end{aligned}
	\end{equation}
	with $k_d^{\text{jf}}$ being a hyperparameter.	
	
	The term $r_s$ rewards a smooth torque profile 
	$r_s = g_s (\bm{\delta}_\tau)$
	with $\bm{\delta}_\tau \in \mathbb{R}^6$ being the differences between the torques applied at two consecutive time steps. 
	
	The term $r_{\theta}$ minimizes the pitch angle, i.e.,
	%
	$r_\theta = g_{\theta}(\theta).$
	%

	Furthermore, similar to the first stage, a bonus is added at the end of one episode.
	For both tasks, we define
	\begin{equation}\small
		\resizebox{0.7\hsize}{!}{$
			\begin{aligned}
				{}^2r_{\text{bonus}}^{\text{jip}} &= \begin{cases}
					0 & \text{not early termination}, \\
					-m^{\text{jip}}\, \overline{h}  & \text{otherwise}.\\
				\end{cases} \\
				{}^2r_{\text{bonus}}^{\text{jf}} &= \begin{cases}
					b^{\text{jf}}\, (\overline{h} + \overline{d}) & \text{not early termination}, \\
					0 & \text{otherwise}.\\
				\end{cases} \\
			\end{aligned}
			$}
	\end{equation}
	with $m^{\text{jip}}$ and $b^{\text{jf}}$ being the hyperparameters.

	

	\section{Evaluation} \label{simulations}
	
	The proposed architecture consistently succeeded in generating effective acrobatic motions, as the one presented in Fig. \ref{fig:results_snapshots} and in the video attachment.
	
	In this section, we report an in-depth analysis of these performance, including ablation studies and comparisons.
	
	\subsection{Technical details}
	We build the PyBullet environment \cite{coumans2019} to emulate Go1's motion. To mimic the parallel spring in the training process, we add the resultant torque by spring extensions (computed by the formula in Appendix~\ref{parallel_elsticity}) to each joint. To run the experiments, we use the RL-Zoo training framework \cite{rl-zoo3}.
	%
	The neural network implementing $\pi_{\mathrm{PPO}}$ is a two-layer DNN with 64 units each. We use $\operatorname{tanh}$ as the activation function and set the learning rate to be $2e^{-4}$. The batch size is 4096, and the discount factor $\gamma$ is 0.999. The clip range is 0.1\footnote{Code is available: \url{https://github.com/francescovezzi/quadruped-springs}}.
	
	\subsection{ES: linear policy vs nonlinear network}\label{comparison_representation}	
\begin{figure}[t]
		\centering
		\setlength{\belowcaptionskip}{-0.cm}	\includegraphics[width=\columnwidth]{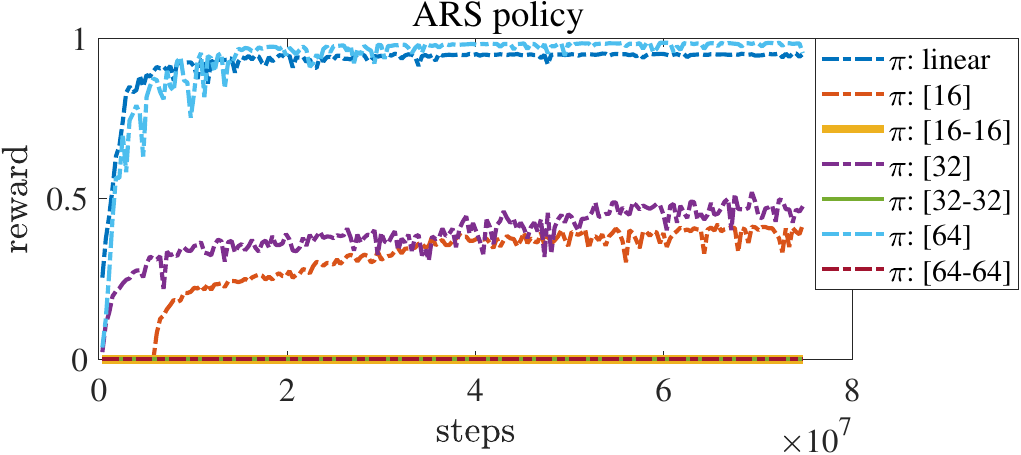}
		\vspace{-0.7cm}
		\caption{Reward profiles for the in-place jumping achieved by the \texttt{ARS} algorithm when using different policy representations.}
		\label{fig:ars_NN}
		\vspace{-0.7cm}
	\end{figure}

	We first compare the linear policy with the NN-based nonlinear policies in \texttt{ARS} training. Taking the in-placed jumping, for example, the learned curves using different representations are plotted in Fig.~\ref{fig:ars_NN}. It demonstrates that the linear policy (see `$\pi\!\!:\!\!\text{linear}$' curve) is among the best representations, obtaining the highest rewards. The comparable performance with higher rewards is achieved by the NN-based policy with one hidden layer of 64 units (`$\pi\!\!:\!\![64]$' curve). Even though, the liner policy converges faster with a smoother reward profile. Thus, it demonstrates that the linear policy representation in the first-stage \texttt{ARS} training process is good enough to achieve competitive results.
	
	\subsection{Learning explosive jumping motion}\label{jumping_motion_results}
\begin{figure*}[t]
		\centering
		\setlength{\belowcaptionskip}{-0.cm}
		\includegraphics[width=2\columnwidth]{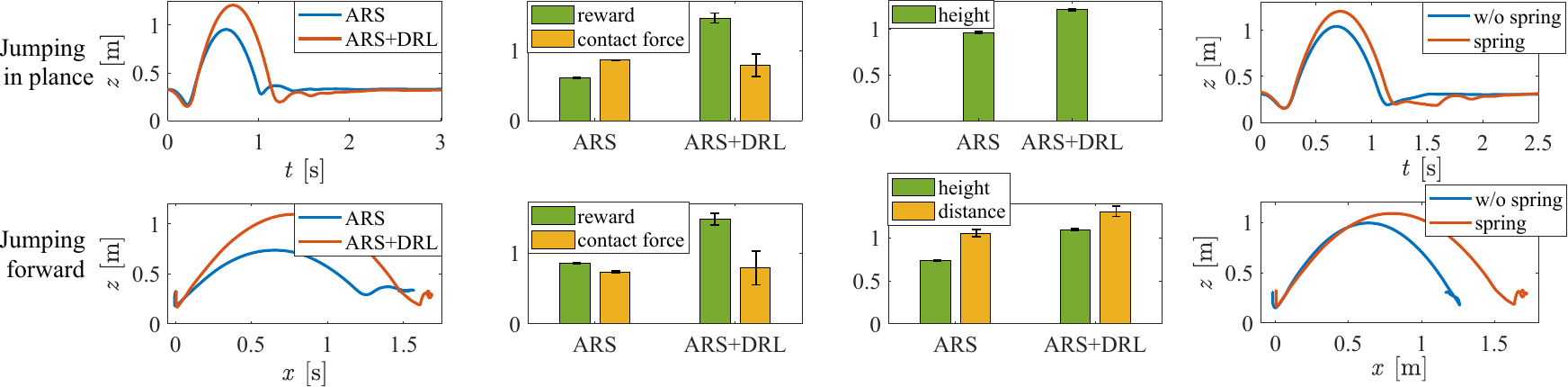}
		\vspace{-0.2cm}
		\caption{Jumping performance comparison. In the second column, a high reward means a better performance, while a low contact force means a compliant landing motion. Note the contact forces are divided by 1000N. In the third column, the `height' and `length' separately denote the maximal jumping height and jumping distance. In the fourth column, the learned CoM trajectories (under `\texttt{ARS+DRL}' policy) for a rigid robot and a soft robot are compared. `w/o spring' denotes the rigid case without springs engaged.}
		\label{fig:jip_jf_results_compare}
		\vspace{-0.2cm}
	\end{figure*}

	Using the hyperparameters in Appendix \ref{parameter_setup_jumping}, the quadrupedal robot learns the jumping motion.  The first three columns in Fig.~\ref{fig:jip_jf_results_compare} present the results under two stages, i.e., `{\texttt{ARS}}' and `\texttt{ARS+DRL}' (described in Fig.~\ref{fig:demonstration_snapshots}).
	
	As can be seen from the first column of Fig.~\ref{fig:jip_jf_results_compare}, both schemes can accomplish two jumping tasks, i.e., jumping in place and jumping forward, successfully. Among the two strategies, `\texttt{ARS+DRL}' achieves the maximal mean reward (see the green bar in the second column), accompanied by the larger jumping height for both tasks (see the green bar in the third column). In addition, compared with `\texttt{ARS}', a longer jumping distance is achieved by `\texttt{ARS+DRL}' in the forward jumping task, as can be seen from the yellow bar in the third column.  Further observations reveal that for in-placed jumping, `\texttt{ARS+DRL}' achieves a smaller contact force, contributing to a compliant landing motion. 
	
	It should be mentioned that the first-stage ES itself (`\texttt{ARS}' in Fig.~\ref{fig:jip_jf_results_compare}) could accomplish two jumping tasks successfully, using the sparse reward defined in Section~\ref{jumping_first_stage}. Considering the `\texttt{ARS+DRL}' policy works better, in the following section, we evaluate it by default.

	\begin{table} 
		\centering
		\vspace{-0.3cm}
		\caption{Jumping performance with/without spring engaged}
		\label{table:elasticity_rigid_robot_jumping}
		\vspace{-0.3cm}
		\setlength{\tabcolsep}{0.8mm}{
			\begin{tabular}{c|c|c|c|c}
				\toprule
				{}& \multicolumn{2}{c|}{Jumping in place}&\multicolumn{2}{c}{Jumping forward}\\
				\cline{2-5}
				{ }&{spring}&{w/o spring}&{spring}&{w/o spring}\\		
				\hline
				{Height [m]}&{\textbf{1.206$\pm$0.010}}&{\textbf{1.045$\pm$0.003}}&{1.100$\pm$0.014}&{0.995$\pm$0.002}\\
				{Distance [m]}&{0.069$\pm$0.015}&{0.207$\pm$0.024}&{\textbf{1.301$\pm$0.073}}&{\textbf{1.057$\pm$0.019}}\\
				{Force [N]}&{799$\pm$189}&{871$\pm$228}&{796$\pm$211}&{800$\pm$369}\\		
				\bottomrule  
				
			\end{tabular}
			\vspace{-0.7cm}
		}
	\end{table}
	
	\subsection{Articulated soft quadruped vs rigid quadruped}
	Using the two-stage learning scheme, we train the control policies for both the articulated soft robot (with parallel springs engaged) and the rigid robot (without engaging parallel springs). To our knowledge, {this is the first report on learning quadrupedal jumping with parallel elasticity in a normal gravity environment}. 
	
	The fourth column of Fig.~\ref{fig:jip_jf_results_compare} compares center of mass (CoM) trajectories in both cases (using `\texttt{ARS+DRL}' policy), and Table~\ref{table:elasticity_rigid_robot_jumping} lists the meaning values of jumping height, jumping distance and the largest landing force. It turns out that a larger height (15.4\% higher) and a longer distance (23.1\% further) are achieved when exploiting the parallel springs. Besides, parallel springs help to gain smaller contact forces with fewer oscillations. More details can be found in the attached video.
	
	\subsection{Beyond single jumping}
\begin{figure}[t]
		\vspace{-0.3cm}
		\centering
		\setlength{\belowcaptionskip}{-0.cm}
		\includegraphics[width=\columnwidth]{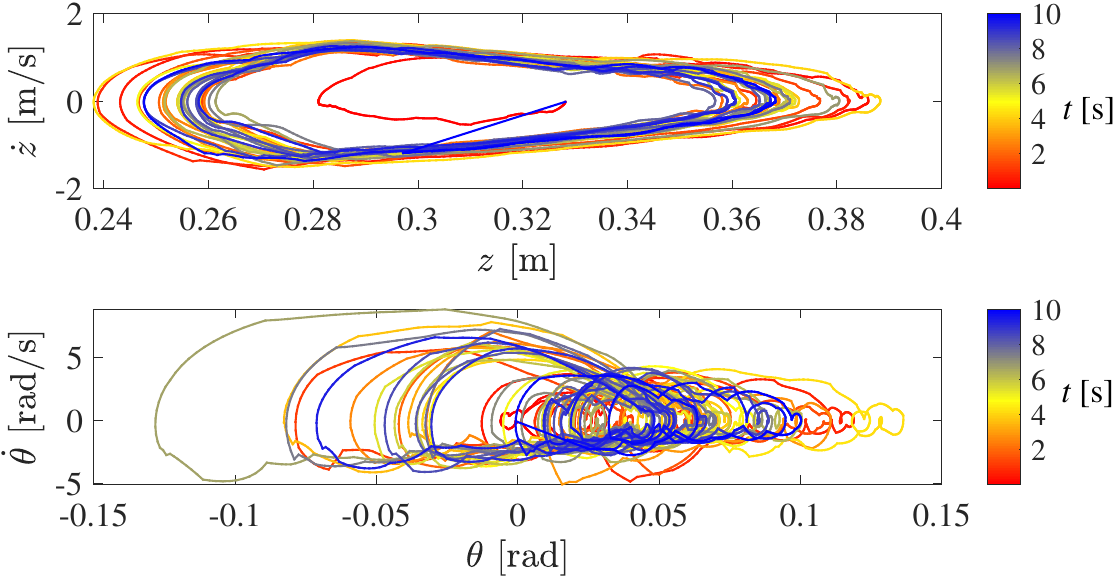}
		\vspace{-0.8cm}
		\caption{{State evolutionary for pronking. Vertical motion (top) and pitch rotation (bottom) converge to the limit cycle.}}
		\label{fig:pronk_state}
		\vspace{-0.3cm}
	\end{figure}

	Using the reward function defined in Appendix~\ref{learning_pronk_reward} and Appendix~\ref{learning_back_reward},  the proposed two-stage learning scheme enables learning the pronking and back-flip motions, which are illustrated in Fig.~\ref{fig:results_snapshots}. Note that no reference motion is used here either. Fig.~\ref{fig:pronk_state} presents the limit cycle behaviour when executing the pronking motion. We find that the vertical CoM and pitch rotation would converge to the limit cycle using the learned policy. Besides, we find that without engaging parallel springs, the back-flip motion is hardly realized.
	
	
	\subsection{Environment randomization}
	To bridge the sim2real gap, we randomize the springs coefficient including damping and stiffness, the total mass, robot mass distribution, CoM offset, and the inertia tensor, as reported in Table~\ref{table：randmize_variable}. In each episode, the environment changes according to the random sampling. 
	
\begin{figure}[t]
		\centering
		\setlength{\belowcaptionskip}{-0.cm}
		\includegraphics[width=\columnwidth]{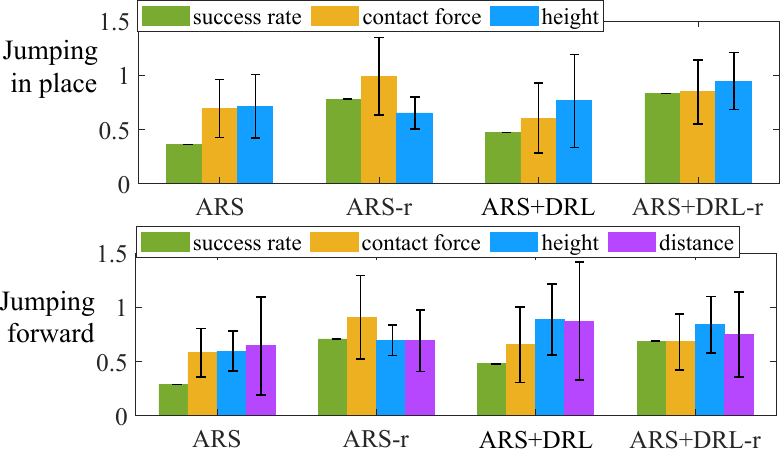}
		\vspace{-0.6cm}
		\caption{Jumping performances in randomized environments. `xx-r' denotes the retrained policy with environment randomization.}
		\label{fig:randommization_compare}
		\vspace{-0.5cm}
	\end{figure}

	To save space, we here only present the results of jumping motion. Fig.~\ref{fig:randommization_compare} compares the jumping performance when using different strategies, where `\texttt{ARS}-r' and `\texttt{ARS+DRL}-r' separately denote the retrained policy using domain randomization techniques \cite{xie2021dynamics}.
	For the in-placed jumping task, domain randomization retraining helps to increase the success rate. In particular, the two-stage retraining with dynamic randomization, i.e., `\texttt{ARS+DRL}-r', increases the jumping height.	For forward jumping, the success rate is also improved by domain randomization. One interesting thing is that the forward jumping distance is decreased compared to the `\texttt{ARS+DRL}' policy without domain randomization. We guess it is because the learning scheme is prone to guarantee the successful rate of forward jumping against environmental uncertainties.

	\begin{table}
		\centering
		\caption{Environment randomization}
		\label{table：randmize_variable}
		\vspace{-0.2cm}
		\setlength{\tabcolsep}{0.8mm}{
			\renewcommand{\arraystretch}{1.2}
			\begin{tabular}{c|c|c|c|c|c}
				\toprule
				\multirow{2}{*}{leg}&{mass}&{CoM}&{trunk}&{spring}&{spring}\\
				{mass}&{offset}&{offset}&{mass}&{stiffness}&{damping}\\				
				\hline
				{$\pm$20\% }&
				{[0 4] [kg]}&{$\pm$[0.2 0 0.2] [m]}&{accordingly}&{$\pm$30\%}&{$\pm$30\%}\\	
				\bottomrule  
			\end{tabular}
		}  
		\vspace{-6mm}
	\end{table}	
	
	\section{Conclusions} \label{conclusions}
	In this work, we propose a two-stage approach for learning highly dynamic quadrupedal motions from scratch. In the first stage, an evolution strategy is used to generate the linear policy, without defining a reference trajectory in advance. Then, we refine and retrain the policy by deep reinforcement learning, achieving higher performance. Compared with current studies that focus on one specific task, it has been demonstrated that, by slightly modifying the reward function, versatile acrobatic motions including pronking and back-flip can be learned, with shared state space and observation space.
	
	Aside from the rigid case, we also apply the method to the articulated soft robot with parallel elasticity. It turned out that the soft robot could still learn highly dynamic motions using the proposed method. Furthermore, exploitation of the parallel springs enhances the motion performance. 
	
	Future work will focus on hardware validation. In addition, we would like to apply the method to humanoid robots, generating general locomotion control polices \cite{ding2020robust,ding2019nonlinear}.
	
	\appendix
	\subsection{Exploiting parallel elasiticity} \label{parallel_elsticity}
	%
	%
	We also explore parallel elasticity in achieving highly dynamic motions by attaching parallel springs to the actuated joints of the Unitree Go1 robot \cite{unitreego1}. For the sake of simplicity, we here consider the springs attached to the joints in the sagittal plane, i.e., thigh and calf joints (see Fig.~\ref{fig:quadrupedal_slip}). The excessive torque on the $i$-th joint caused by the spring deformation ($\tau^\text{s}$) is computed as 
	$\tau^\text{s}_i = k_i \left( q_i^{\text{ref}} - q^\text{0}_i \right) + c_i  \dot{q}_i^{\text{ref}} $
	with $k_i \in \mathbb{R} $ and $c_i \in \mathbb{R} $ separately being the spring stiffness and damping, $q_i^{\text{ref}}  \in \mathbb{R}$, $\dot{q}_i^{\text{ref}}  \in \mathbb{R} $, and $q^\text{0}_i \in \mathbb{R} $ separately denoting the commanded angle, angular velocity) and rest angle of the $i$-th joint.

	\subsection{Rewards for pronking and back-flip}\label{reward_all}
	
	\subsubsection{Reward for pronking} \label{learning_pronk_reward}
	
	
	To begin with, we evaluate single jump by
	\begin{equation}
		p_{\text{sj}} = w_h h_n + w_d d_n.
	\end{equation}
	
	Then, we define the performance array for $n$ jumps during one episode as 
	$p_j = \left[p_{\text{sj}}^1, p_{\text{sj}}^2, \dots, p_{\text{sj}}^n  \right]$.
	%
	Given $p_j$, we can compute the average performance ${p}_j^{\text{avg}}$ and the performance entropy $S(p_j)$ (considering $p_{\text{sj}}^i \in \left[0, 1\right] $ and $\sum_{i = 1}^{n}p_{\text{sj}}^i=1$). The sparse reward used for the first-stage learning is then defined as
	\begin{equation}
		\resizebox{0.7\hsize}{!}{$
			\begin{aligned}
				&r_{\text{avg}} = {p}_j^{\text{avg}} \left\{
				g_\theta(\overline{\theta}) + w_t t_n + w_s S(p_j) + k
				\right\}, \\
				&{}^1r^{\text{pronking}} = w_{\text{avg}} r_{\text{avg}} + w_{\text{max}} p_{\text{sj}}^{\max} +w_c c_j + b,
			\end{aligned}
			$}
	\end{equation}
	where $c_j$ is the number of jumps whose performance is above a certain threshold, $b$ is a bonus for a successful termination, $t_n$ is the ratio between the time the episode ends and the maximal episode time, $k$ is a constant and $p_{\text{sj}}^{\max}$ is the maximal element of the array $p_j$. $w_{\times}$ denotes the hyperparameter.
	
	For the second-stage training, the reward function is similar to the one already defined in \eqref{eq:reward_second_stage}, added by two additional terms. One term ($r_p$) is used to minimize the actuators' energy consumption, and the other term ($r_j$) provides a reward each time the robot successfully performs a jump, which are separately defined as
	\begin{equation}
		\resizebox{0.7\hsize}{!}{$
			\begin{aligned}
				&r_p = g_p\left(E\right), \quad
				&r_j = w_j {p}_j^{\text{avg}} \left[ g_s\left(S\left(p_j\right)\right) + l_j\right],
			\end{aligned} 
			$}
	\end{equation}
	where $E$ is the energy cost in one episode, $l_j$ is a constant reward, and the weight $w_j$ is $0$ if the jump performance is below a certain threshold. 

	\subsubsection{Reward for back-flipping}\label{learning_back_reward}
	
	
	Denoting $\theta_n \in \mathbb{R}$ as the normalized maximum pitch angle in one episode (computed by \eqref{normalization_xx}), the reward function for the first-stage \texttt{ARS} training is defined as
	\begin{equation}
		{}^1r^{\text{flip}} = w_h \overline{h} + w_\theta \theta_n + w_{\text{h}\theta} (\overline{h} \theta_n) + b,
	\end{equation}
	where $b$ is a bonus if the episode ends successfully, $w_{\times}$ denotes the hyperparameter.
	
	The reward for the second-stage DRL training is similar to the one defined in \eqref{eq:reward_second_stage}. One exception is the reward with regard to the jumping distance ($r_d$ in \eqref{eq:reward_second_stage}) is discarded and the one with regard to the pitch angle is proportional to it, i.e.,
	%
	$r_\theta = g_{\theta}(\theta)$. Another point is that we change $r_{\text{bonus}}$ in \eqref{eq:reward_second_stage} to be
	\begin{equation}
		\resizebox{0.7\hsize}{!}{$
			r_{\text{bonus}} = \begin{cases}
				w_{\text{bonus}} (\overline{h} \theta_n)  & \text{not early termination}, \\
				0 & \text{otherwise}. 
			\end{cases}
			\label{fig:reward_function_backflip}
			$}
	\end{equation}	
	
	\subsection{Hyperparameters for learning jumping motion} \label{parameter_setup_jumping}
	
	Hyperparameters for the first-stage and second-stage training are separately listed in Table~\ref{table:rew_hp_ars} and Table~\ref{table:reward_hp_ppo}.


	\begin{table}
		\centering
		\caption{Reward penalties for the first-stage \texttt{ARS} training.}
		\label{table:rew_hp_ars}
		\vspace{-0.3cm}
		\setlength{\tabcolsep}{0.9mm}{
			\renewcommand{\arraystretch}{1.4}
			\begin{tabular}{c|c|c|c|c|c|c|c|c|c}
				\toprule
				\multirow{2}{*}{Jumping}&{$h_f^\text{jip}$}&{$a_\theta^\text{jip}$}&{$b_\theta^\text{jip}$}&{$a_d^\text{jip}$}&{$b_d^\text{jip}$}&{$c_h^\text{jip}$}&{$b^\text{jip}$}&{$q^\text{jip}$}&{$m^\text{jip}$}\\	
				{in place}&{0.9}&{0.3}&{0.0225}&{0.05}&{0.05}&{0.7}&{0.1}&{0.08}&{0.064}\\
				\hline
				\multirow{2}{*}{Jumping}&{$h_f^\text{jf}$}&{$a_\theta^\text{jf}$}&{$b_\theta^\text{jf}$}&{$d_f^\text{jf}$}&{$c_d^\text{jf}$}&{$c_h^\text{jf}$}&{$b^\text{jf}$}&{$q^\text{jf}$}&{$m^\text{jf}$}\\	
				{forward}&{0.3}&{0.25}&{0.0225}&{1.3}&{0.5}&{25}&{0.1}&{0.08}&{0.096}\\			
				\bottomrule  
			\end{tabular}
		}  
	\end{table}

	\begin{table}
		\vspace{-2mm}
		\centering
		\caption{Reward penalties for the second-stage \texttt{PPO} training.}
		\label{table:reward_hp_ppo}
		\vspace{-0.3cm}
		\setlength{\tabcolsep}{0.7mm}{
			\renewcommand{\arraystretch}{1.4}
			\begin{tabular}{c|c|c|c|c|c|c}
				\toprule
				\multirow{4}{*}{Jumping}&{$a_h^\text{jip}$}&{$h_{\min}^\text{jip}$}&{$h_{\max}^\text{jip}$}&{$a_c^\text{jip}$}&{$f_{\min}^\text{jip}$}&{$a_d^\text{jip}$}\\	
				{}&{0.01}&{0.29}&{1.3}&{$1.5e^{-4}$}&{800}&{$6.5e^{-4}$}\\
				\cline{2-7}
				{in place}&{$b_d^\text{jip}$}&{$a_s^\text{jip}$}&{$b_s^\text{jip}$}&{$a_{\theta}^\text{jip}$}&{$b_{\theta}^\text{jip}$}&{$m^\text{jip}$}\\	
				{}&{40}&{$3e^{-3}$}&{0.1}&{$4.2e^{-3}$}&{26}&{0.25}\\
				\hline
				\multirow{4}{*}{Jumping}&{$a_h^\text{jf}$}&{$h_{\min}^\text{jf}$}&{$h_{\max}^\text{jf}$}&{$a_c^\text{jf}$}&{$f_{\min}^\text{jf}$}&{$k_d^\text{jf}$}\\	
				{}&{$6.5e^{-3}$}&{0.29}&{1.1}&{$1.2e^{-4}$}&{800}&{$1.52e^{-2}$}\\
				\cline{2-7}
				{forward}&{$d_{\max}^\text{jf}$}&{$a_s^\text{jf}$}&{$b_s^\text{jf}$}&{$a_{\theta}^\text{jf}$}&{$b_{\theta}^\text{jf}$}&{$b^\text{jf}$}\\	
				{}&{1.3}&{$3e^{-3}$}&{0.1}&{$4.2e^{-3}$}&{26}&{0.025}\\						
				\bottomrule  
			\end{tabular}
		}  
	\end{table}

	\clearpage
	%
	\balance

	%

\end{document}